\documentclass[runningheads]{llncs}
\usepackage{graphicx}
\usepackage{algorithmic}
\usepackage{amssymb}
\usepackage{algorithm}
\usepackage{color}
\usepackage{siunitx}
\usepackage{booktabs}
\usepackage{threeparttable}
\usepackage{multirow}
\usepackage{marvosym}
\usepackage{amsmath}

\definecolor{gray}{RGB}{128, 128, 128}
\definecolor{red}{RGB}{255, 0, 0}
\begin{document}
\title{Seg4Reg+: Consistency Learning between Spine Segmentation and Cobb Angle Regression
% \thanks{Supported by organization x.}
}
\titlerunning{Seg4Reg+}
% \author{Anonymous MICCAI 2021 submission}
% \institute{Paper ID 1599\\
% \email{**@******.***}}
\author{%
Yi Lin\inst{1}\textsuperscript{\Letter} \and
Luyan Liu\inst{1} \and
Kai Ma\inst{1} \and
Yefeng Zheng\inst{1}}

\authorrunning{Yi Lin et al.}
\institute{Tencent Jarvis Lab, Shenzhen, China\\
\email{linyi.pk@gmail.com}}
\maketitle              % typeset the header of the contribution
\begin{abstract}
Automated methods for Cobb angle estimation are of high demand for scoliosis assessment. Existing methods typically calculate the Cobb angle from landmark estimation, or simply combine the low-level task (e.g., landmark detection and spine segmentation) with the Cobb angle regression task, without fully exploring the benefits from each other. In this study, we propose a novel multi-task framework, named Seg4Reg+, which jointly optimizes the segmentation and regression networks. We thoroughly investigate both local and global consistency and knowledge transfer between each other. Specifically, we propose an attention regularization module leveraging class activation maps (CAMs) from image-segmentation pairs to discover additional supervision in the regression network, and the CAMs can serve as a region-of-interest enhancement gate to facilitate the segmentation task in turn. Meanwhile, we design a novel triangle consistency learning to train the two networks jointly for global optimization. The evaluations performed on the public AASCE Challenge dataset demonstrate the effectiveness of each module and superior performance of our model to the state-of-the-art methods.
\end{abstract}
\section{Introduction}
Adolescent idiopathic scoliosis (AIS) causes a structural, lateral, rotated curvature of the spine that arises in children at or around puberty~\cite{weinstein2008adolescent}. The Cobb angle, derived from a anterior-posterior X-ray and measured by selecting the most tilted vertebra, is the primary means for clinical diagnosing of AIS. However, manual measurement of the Cobb angle requires professional radiologists to carefully identify each vertebra and measure angle, which is time-consuming and could suffer from a large inter-/intra-observer variety. Hence, it is needed to provide an accurate and robust method for quantitative measurement of Cobb angle automatically.

Numerous computer-aided methods have been developed for automated Cobb angle estimation. Conventional methods utilized active contour model~\cite{anitha2012automatic}, customized filter~\cite{anitha2014automatic} and charged-particle models~\cite{sardjono2013automatic} for spine segmentation to calculate the Cobb angle, which are computationally expensive and the unclear spine boundary will result in inaccurate estimations. Recently, deep learning based methods~\cite{chen2019automated,kim2020automation,sun2017direct,wu2017automatic,yi2020vertebra,zhang2021automated} have been proposed to consolidate the tasks of vertebral landmark detection with Cobb angle estimation to improve the robustness of spinal curvature assessment. And Seg4Reg~\cite{lin2019seg4reg}, which won the 1st place in the AASCE\footnote[1]{https://aasce19.grand-challenge.org} challenge, regarded the segmentation results of spine as the input of the regression network for Cobb angle estimation. Its superior performance owns to the segmentation mask of spine, which retains the shape information and filters the distractions (e.g., artifacts and local contrast variation). However, the performance of this method depends heavily on the segmentation results. 

Although the above methods have achieved great success, their applications to Cobb angle estimation suffer from three limitations: 1) The methods~\cite{chen2019automated,kim2020automation,sun2017direct,wu2017automatic,yi2020vertebra,zhang2021automated} relying on the landmark coordinates are susceptible since a small error in landmark coordinates may cause a huge mistake in angle estimation; 2) the two-stage frameworks~\cite{chen2019automated,kim2020automation,lin2019seg4reg,zhang2021automated} often suffers from the error accumulation; and 3) the %separately trained
cascaded networks\cite{chen2019automated,kim2020automation,lin2019seg4reg} cannot guarantee a global optimum.

\begin{figure}[!t]
    \centering
    \includegraphics[width=\textwidth]{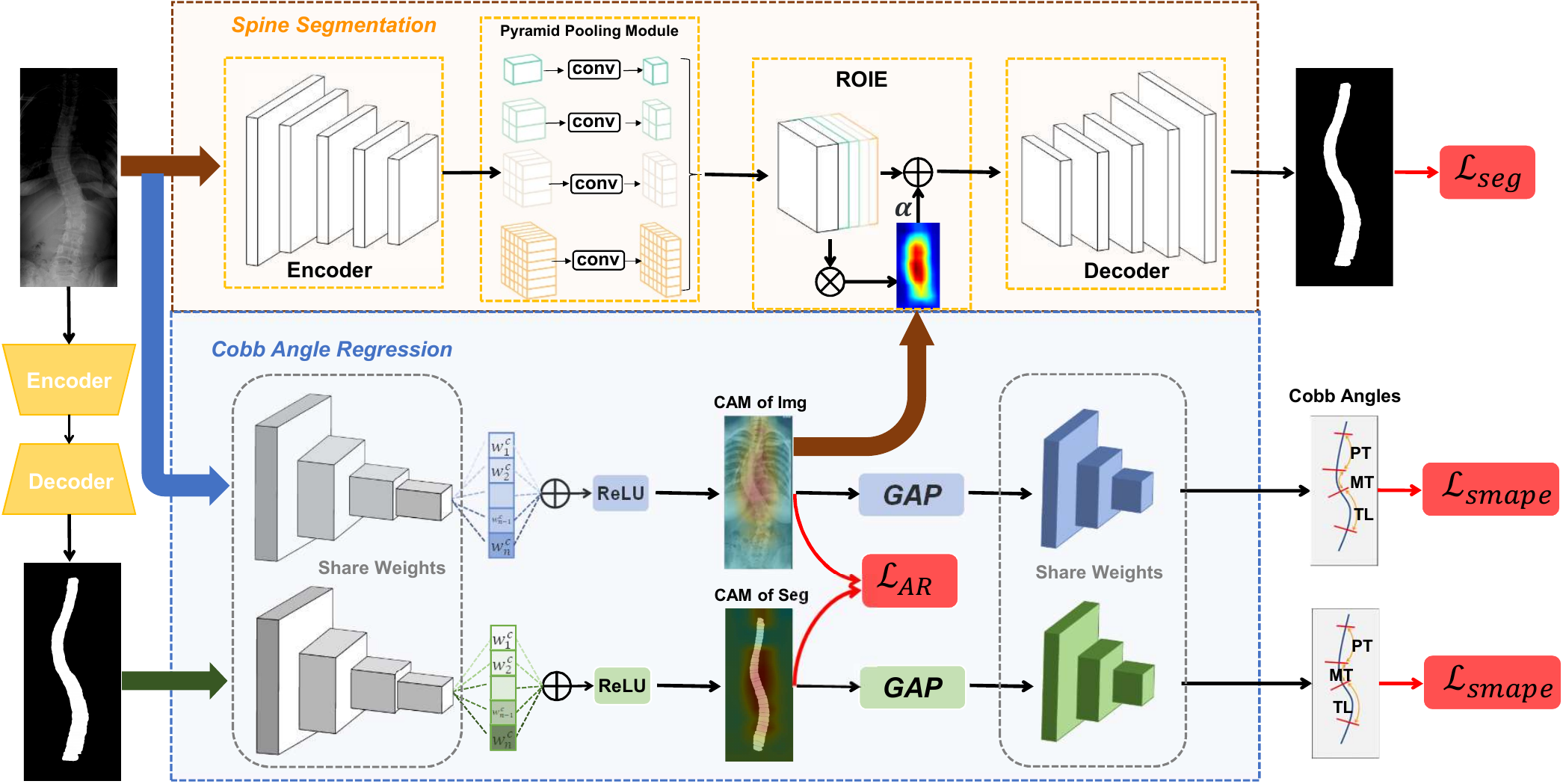}
    \caption{An overview of the proposed framework.}
    \label{fig_framework}
\end{figure}

In this paper, we propose a novel consistency learning framework, named Seg4Reg+, which incorporates segmentation into the regression task, as shown in Fig.~\ref{fig_framework}. The segmentation task extracts representative features for the regression task by an \textbf{attention regularization} (AR) module with auxiliary constraint on the class activation map (CAM). And the regression task is able to provide specific hints for the segmentation task by a \textbf{region-of-interest enhancement} (ROIE) gate to force the segmentation network to pay more attention to the important area. To reach the global optimum, we further design a novel \textbf{triangle consistency learning} scheme for end-to-end training. In summary, our main contributions are as follows:
\begin{itemize}
    \item We propose a novel consistency learning framework, named Seg4Reg+, incorporating segmentation and regression tasks with an AR module and ROIE gate to boost the performance of both tasks.
    \item We design a triangle consistency learning scheme for end-to-end training.
    \item Extensive evaluations on the public AASCE Challenge dataset demonstrate the effectiveness of each module and superior performance of our model to the state-of-the-art methods.%\textit{We will release the source code once the paper is accepted for publication.}
\end{itemize}

\section{Method}

As illustrated in Figure~\ref{fig_framework}, the proposed Seg4Reg+ model consists of a segmentation network $N_S$ and a regression network $N_R$. We first pre-train the two networks separately for approximately optimized results to speed up the training process. Then the two networks are boosted by each other by the ROIE gate and AR module. In addition, a novel training strategy named triangle consistency learning scheme is designed for end-to-end training. In the following, we first introduce the proposed ROIE gate and AR module, then illustrate the triangle consistency learning scheme.

\noindent\textbf{ROIE Gate.}
We first train $N_S$ to roughly segment the spine region. We adopt the same network as Seg4Reg~\cite{lin2019seg4reg} for a fair comparison. Specifically, we modify the PSPNet~\cite{zhao2017pyramid} by replacing the pooling layer with the dilated convolution in the pyramid pooling module and take ResNet-50~\cite{he2016deep} as the backbone. The objective function is the weighted Dice loss and cross-entropy loss:
\begin{equation}
     \mathcal{L}_{\mathrm{seg}}(s(x), y) = \sum_i\left(1 - \frac{2\times s(x_i)y_i}{s(x_i) + y_i}\right) + \lambda\sum_i(-y_i\log(s(x_i))),
     \label{eq1}
\end{equation}
where $x$ denotes the input image, and $s(x_i)$ and $y_i$ denote the label of pixel $x_i$ of prediction and ground truth, respectively, and $\lambda$ denotes the hyperparameter weighting the two losses.

To boost the performance of $N_S$ with $N_R$, the ROIE gate is designed as an attention mechanism to transfer the specific hints from $N_R$ to $N_S$. The proposed ROIE gate is inspired by CAM, which is the most common technique in weakly supervised segmentation methods~\cite{zhou2016learning}. We expect that the CAM of $N_R$ can incorporate the refined prior information about the spine area into the segmentation process. In addition, the value of each pixel on the CAM represents its significance to the regression output, which in turn guides $N_S$ to pay more attention to the important areas (i.e., the most tilted vertebra endplates). Specifically, we treat CAM of $N_R$ as attention map, and perform a matrix multiplication between CAM and the feature map $f_m(x)$ from the middle layer of $N_S$. Then, we multiply the result by a scalar parameter $\alpha$ and perform an element-wise sum operation with the feature $f_m(x)$ to obtain the final output $f'_m(x)$ as follows:
\begin{equation}
    f'_m(x) = \alpha(C(r(x))\circ f_m(x)) + f_m(x),
    \label{eq4}
\end{equation}
where $C(\cdot)$ is a CAM that indicates the discriminative part with respect to the regression results, $\alpha$ is a learnable parameter which is initialized as 0, $\circ$ denotes multiplication function. It can be inferred from Equation~(\ref{eq4}) that the resulting feature $f'_m(x)$ combines global contextual view and selectively aggregates contexts according to the CAM, thus improving intra-class compact and semantic consistency.

% \subsubsection{AR Module.}
\noindent\textbf{AR Module.}
For $N_r$, we modify the state-of-the-art classification network by replacing the last convolutional layer with the output channel corresponding to three clinically relevant Cobb angles: proximal thoracic (PT), main thoracic (MT) and thoracolumbar/lumbar (TL). And the activation function in the last layer is set to the sigmoid function. Here, we design a novel objective function based on symmetric mean absolute percentage error, named SMAPE loss:
\begin{equation}
\label{eq2}
    \mathcal{L}_{\mathrm{SMAPE}}(r(x), \hat{y}) = \frac{\sum_{i=1}^n|\hat{y}_i - r(x_i)|}{\sum_{i=1}^n|\hat{y_i} + r(x_i) + \epsilon|},
\end{equation}
where $\hat{y_i}$ and $r(x_i)$ denote the ground truth and prediction of $i$th angle of the total $n=3$ Cobb angles (i.e., PT, MT and TL), and $\epsilon$ is a smooth factor.

To boost the performance of $N_R$ with $N_S$, the AR module is designed to explore the hidden state representation of the $N_R$ via classification activation mapping (CAM) to force it to focus on the spine area. Specifically, to integrate regularization on $N_R$, we expand the $N_R$ into a shared-weight Siamese structure. One branch takes the concatenation of the raw image and its corresponding segmentation mask as input, and the other directly takes the segmentation as input. The output activation maps from two branches are regularized by mean absolute error to guarantee the consistency of CAMs, and the regression network in consequence is forced to focus on the spine area. The objective function is:
\begin{equation}
\label{eq3}
    \mathcal{L}_{\mathrm{AR}} = \| C(x, s(x)) - C(s(x))\|_1,
\end{equation}

\begin{algorithm}[!ht]
\caption{Triangle Consistency Learning.}
\label{algo1}
\begin{algorithmic}[1]
\REQUIRE ~~\\ 
Input image, $x\in X$;\\
Ground truth of segmentation mask of spine, $y\in Y$;\\
Ground truth of three Cobb angles (PT, MT and TL), $\hat{y}\in \hat{Y}$;\\
\ENSURE ~~\\ 
% Dual-task model’s parameter $\theta_1$ for segmentation network, $\theta_1$ for cobb angle regression network;\\
$s(\cdot)$: Segmentation task of network $N_S$ with parameter $\theta_1$;\\
$r(\cdot)$: Regression task of network $N_R$ with parameter $\theta_2$;\\
$C(\cdot)$: Class activation map (CAM) generated by $N_R$;\\
\textit{\textcolor{gray}{\textbackslash\textbackslash Training basic $N_S$.}}
\WHILE{stopping criterion not met}
\STATE Compute the segmentation loss $\mathcal{L}_{\mathrm{seg}}(s(x), y)$ with Equation~(\ref{eq1});\\ %$\mathcal{L}_{seg}(f(x), s) = (1 - \frac{2\times\sum_i^{n}f(x_i)s_i}{\sum_i^{n}f(x_i) + \sum_i^{n}s_i}) + \lambda (-\frac{1}{n} \sum_i^{n} s_i log(f(x_i)))$\
\STATE Update parameters $\theta_1$ of $N_S$ by backpropagation;
\ENDWHILE

\textit{\textcolor{gray}{\textbackslash\textbackslash Training $N_R$ with AR.}}
\WHILE{stopping criterion not met}
\STATE Compute the SMAPE loss $\mathcal{L}_{\mathrm{SMAPE}}(r(x, s(x)), \hat{y})$ with Equation~(\ref{eq2});\\ 
\textit{\textcolor{gray}{\textbackslash \textbackslash$(x, s(x))$ means concatenation of raw image and its segmentation mask.}}\\
\STATE Compute the SMAPE loss $\mathcal{L}_{\mathrm{SMAPE}}(r(s(x)), \hat{y})$ with Equation~(\ref{eq2});\\
\STATE Compute the attention regularization loss $\mathcal{L}_{\mathrm{AR}} = \| C(x, s(x)) - C(s(x))\|_1$; \\
\STATE Update the regression network parameters $\theta_2$;
\ENDWHILE

\textit{\textcolor{gray}{\textbackslash\textbackslash Fine-tuning $N_S$ with ROIE.}}
\WHILE{stopping criterion not met}
\STATE Add local consistency constraints to $N_S$ with Equation~(\ref{eq4});\\
% \STATE Compute the regression loss $L_{smape}(g(s), g(f(x, C(x))))$\\
% \STATE freeze the $\theta_2$ and update the $\theta_1$
\STATE Update the parameters $\theta_1$ of $N_S$;\\
\ENDWHILE

\textit{\textcolor{gray}{\textbackslash\textbackslash Fine-tuning $N_S$ by SMAPE loss.}}
\WHILE{stopping criterion not met}
% \STATE Add local consistency constraints to segmentation network with Equation~\ref{eq4};\\
\STATE Compute the regression loss $\mathcal{L}_{\mathrm{SMAPE}}(r(s), r(s(x, C(x))))$;\\
\STATE Freeze the $\theta_2$ and update the $\theta_1$;\\
\ENDWHILE

\textit{\textcolor{gray}{\textbackslash\textbackslash Fine-tuning $N_R$ with refined segmentation.}}
\STATE Repeat steps 5-10
\RETURN $\theta_1$ and $\theta_2$
\end{algorithmic}
\end{algorithm}

\begin{figure}[!t]
    \centering
    \includegraphics[width=\textwidth]{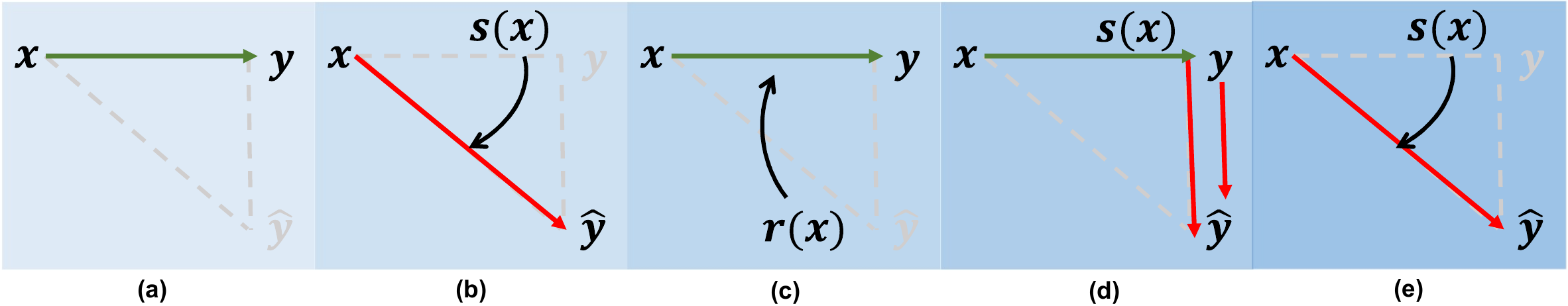}
    \caption{The training strategy of triangle consistency learning: (a) training $N_S$, (b) training $N_R$ with AR, (c) fine-tuning $N_S$ with ROIE, (d) fine-tuning $N_S$ by the SMAPE loss, and (e) fine-tuning $N_R$ with refined segmentation.} 
    % The green line and red line denote the segmentation and regression task, respectively.}
    \label{fig_trainstrategy}
\end{figure}
\noindent\textbf{Triangle Consistency Learning.}
Inspired by the inference-path invariance theory~\cite{zamir2020robust} which declares that inference paths with the same endpoints, but different intermediate domains, yield similar results. The segmentation process is essentially an auxiliary task for the regression task, thus the regression network has the potential to optimize the segmentation network. Based on this assumption, we design a novel training strategy named as triangle consistency learning for end-to-end training, are shown in Figure~\ref{fig_trainstrategy}. The details of the proposed training strategy are shown in Algorithm~\ref{algo1}, which can be divided into five processes: 1) training basic $N_S$ (steps 1-4); 2) training $N_R$ with AR (steps 5-10); 3) fine-tuning $N_S$ with ROIE (steps 11-14); 4) fine-tuning $N_S$ with SMAPE loss on the regression output of $s(x)$ and $y$ (steps 15-18); and 5) fine-tuning $N_R$ with refined segmentation (step 19). In the fourth process, we compute the SMAPE loss for the regression outputs of segmentation results and the corresponding ground truth, denoted $r(s(x))$ and $r(y)$, respectively. Then we freeze the parameters of $N_R$ and optimize $N_S$ by the backpropagation. In this way, we can generate more suitable segmentation mask for $N_R$, superior to traditional segmentation. 

\section{Experiments}
\noindent\textbf{Data and Implementation Details.}
We use the public dataset of MICCAI 2019 AASCE Challenge~\cite{wu2017automatic}, which consists of 609 spinal anterior-posterior X-ray images to evaluate our method. The dataset is divided by the provider into 481 images for training and 128 images for testing. We evaluate the proposed method using two metrics, symmetric mean absolute percent error (SMAPE) and mean absolute error (MAE). And we evaluate the segmentation results using five performance metrics including the Jaccard index (JA), Dice coefficient (Dice), pixel-wise accuracy (pixel-AC), pixel-wise sensitivity (pixel-SE), and pixel-wise specificity (pixel-SP).

We pre-process the data before inputting it into our network. First, we resize the image to [512, 256]. Then, we linearly transform the Cobb angles into [0, 1], and augment the dataset by randomly flipping, rotating (\ang{-45}, \ang{45}), and rescaling with the factor between (0.85, 1.25). We train the $N_S$ for 90 epochs using ADAM optimization with learning rate $1\times10^{-4}$ and weight decay $1\times10^{-5}$. And for regression, we train the network for 200 epochs in total with learning rate $1\times10^{-3}$ and weight decay $1\times10^{-5}$.

\begin{table}[!t]
  \centering
  \caption{The ablation study for each part of Seg4Reg+. AR: attention regularization module, ROIE: region-of-interest enhancement, TCL: triangle consistency learning, Img: raw image as input, and Seg: segmentation mask.}
  \label{tab1}
  \setlength{\tabcolsep}{2mm}{
    \begin{tabular}{cccccc|cc}
    \toprule
    Baseline & AR & ROIE & TCL & Img & Seg & MAE & SMAPE (\%)\\
    \midrule 
    \checkmark &&&& \checkmark && 6.34, 7.77, 8.01 & 12.32 \\
    \checkmark & \checkmark &&& \checkmark && 4.55, 5.75, 5.92 &  9.39\\
    \checkmark & \checkmark & \checkmark && \checkmark && 4.21, 5.32, \textbf{5.22}  & 9.32 \\
    \checkmark & \checkmark & \checkmark & \checkmark & \checkmark &&  \textbf{4.01, 5.16}, 5.51 &  \textbf{9.17}\\
    \hline
    \checkmark &&&&& \checkmark & 6.51, 6.22, 7.17 & 10.95 \\
    \checkmark & \checkmark &&&& \checkmark & 4.03, 6.38, 5.80 &  9.52\\
    \checkmark & \checkmark & \checkmark &&& \checkmark &\textbf{3.61}, 4.90, 5.53 & 9.01 \\
    \checkmark & \checkmark & \checkmark & \checkmark && \checkmark &  5.13, \textbf{4.73, 5.24} & \textbf{8.92}\\
    \hline
    \checkmark & \checkmark & \checkmark & \checkmark & \checkmark & \checkmark &  \textbf{3.88, 4.62, 4.99} & \textbf{8.47}\\
    \bottomrule
    \end{tabular}}
\end{table}

\noindent\textbf{Ablation Studies.}
To verify the effectiveness of each module in the proposd Seg4Reg+ approach, we conduct a set of experiments for ablation study, which are shown in Table~\ref{tab1}. We first test with the raw image as input, the AR module has a 2.93\% improvement in SMAPE compared to baseline, which validates that the AR module can gain from segmentation and benefit for the regression task. By combining the ROIE gate, the performance can be further improved by 0.07\% (the improvement with the segmentation mask as input is more significant with 0.51\% boost in SMAPE), which demonstrates $N_R$ can boost $N_S$ in turn. And applying triangle consistency learning could improve the SMAPE by 0.15\%.
Similar trend can also be observed when we test with segmentation results as input. Finally, when we concatenate the raw image and segmentation mask together as input, the performance can by further improved up to 8.47\%. 

\begin{figure}[!t]
    \centering
    \includegraphics[width=\textwidth]{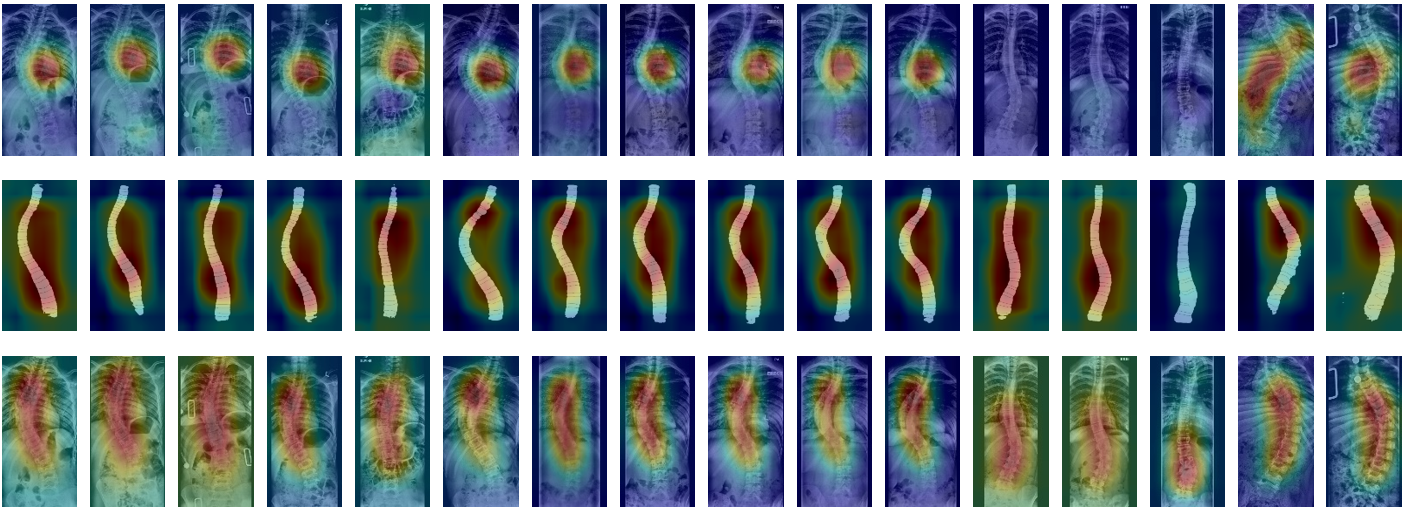}
    \caption{The visualization of CAMs with and without the AR module. The top and middle rows show the result of baseline method (without AR) using raw image and segmentation mask as input, respectively. And the bottom row shows the CAMs generated from the baseline method with AR.}
    \label{fig_AR}
\end{figure}
\begin{figure}[!t]
    \centering
    \includegraphics[width=\textwidth]{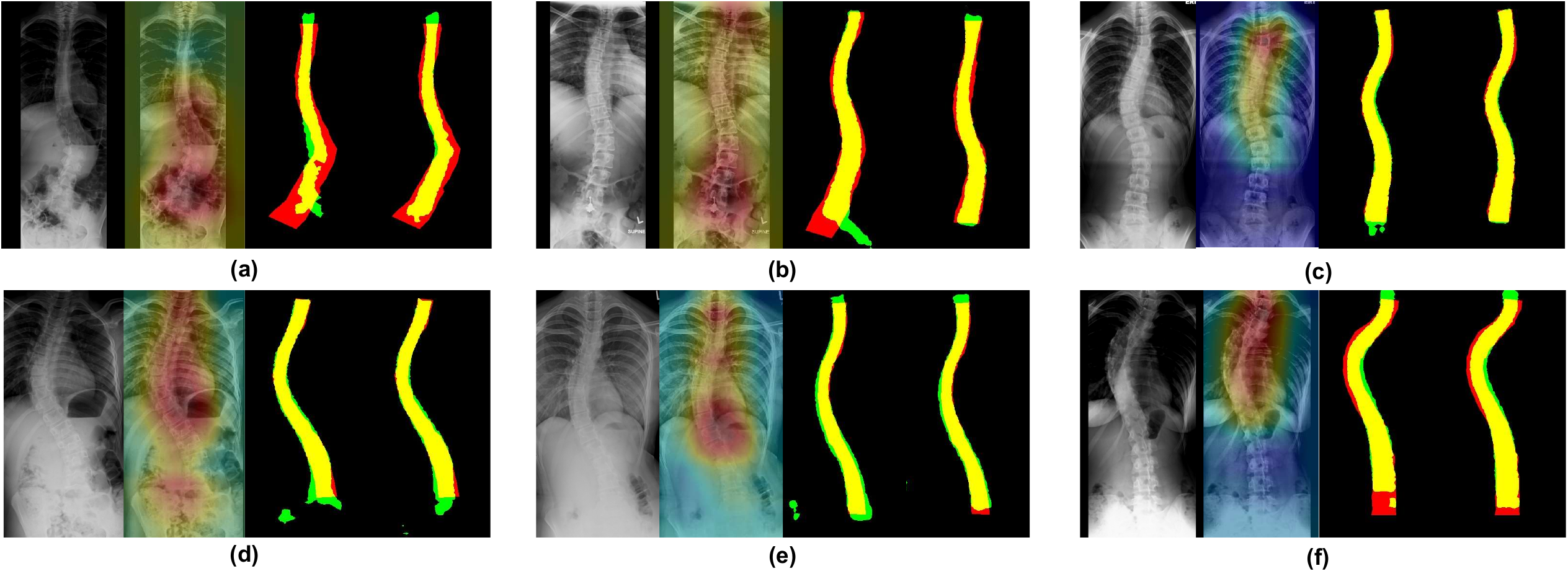}
    \caption{Examples of segmentation results. Each group shows the original image, CAM, and segmentations without and with ROIE, respectively. The yellow mask is true positive, red mask is false negative, and green mask is false positive.}
    \label{fig2}
\end{figure}

\noindent\textbf{Effectiveness of AR Module.}
Figure~\ref{fig_AR} shows the results of the baseline method and the AR module. From the first row in Figure~\ref{fig_AR}, we can see several drawbacks of the CAM generated by the baseline method: 1)  focusing only on the local region (e.g., columns 1-11), 2) the relative inferior ability of feature extraction (e.g., columns 12-14), and 3) the vulnerability to the blurred images (e.g., columns 15-16). And the bottom row in Figure~\ref{fig_AR} shows that with the AR module, $N_R$ can make more precise prediction with more reasonable perspective view. Particularly, in column 14, our AR module predicts more precise attention area while the baseline method generates very weak attention maps.

\begin{table}[!t]
    \caption{Comparison of segmentation performance (\%) of ROIE with other methods.}
    \centering
    \setlength{\tabcolsep}{3.5mm}{
    \begin{tabular}{c|ccccc}
        \toprule
        Methods & JA & Dice & pixel-AC & pixel-SE & pixel-SP \\
        \hline
        w/o CAMs & 75.47 & 86.02 & 94.83 & 86.27 & 98.21\\
        MDC~\cite{wei2018revisiting}      & 75.91 & 86.31 & \textbf{95.73} & \textbf{89.08} & 98.12 \\
        D-Net~\cite{hong2015decoupled}    & 76.99 & 87.00 & 95.02 & 86.61 & 98.46\\
        MB-DCNN~\cite{xie2020mutual}  & 76.17 & 86.47 & 95.12 & 87.18 & 98.19\\
        Ours     & \textbf{77.86} & \textbf{87.55} & 95.49 & 88.06 & \textbf{98.42}\\
        \bottomrule
    \end{tabular}
    }
    \label{tab2}
\end{table}
% \subsubsection{Effectiveness of ROIE Gate.}
\noindent\textbf{Effectiveness of ROIE Gate.}
We evaluate both qualitative and quantitative segmentation results of our ROIE module. Table~\ref{tab2} shows that applying the ROIE gate to the origin segmentation model promotes JA from 75.47\% to 77.86\%, and the visualization results in Figure~\ref{fig2} show that the ROIE gate is helpful for false positive reduction. Furthermore, we compare different ways of transferring CAM to the segmentation network, i.e., MDC~\cite{wei2018revisiting}, D-Net~\cite{hong2015decoupled} and MB-DCNN~\cite{xie2020mutual}. 
However, simply delivering CAMs to $N_S$, these methods would be affected by inaccurate CAMs. And our ROIE gate alleviates this problem by fusing the CAM adaptively with a learnable parameter. For a fair comparison, all of these methods use the same segmentation architecture. It shows that our ROIE gate improves JA by 0.9\% and 1.7\% over D-Net and MB-DCNN, respectively.

\begin{table}[!t]
    \caption{Comparison of regression performance with the state-of-the-art methods.}
    \small
    \centering
    \setlength{\tabcolsep}{1.5pt}{
    \begin{tabular}{c|c|c|c|c|c|c|c|c|c|c}
        \toprule
        Methods & \scriptsize A-Net~\cite{chen2019automated} & \scriptsize L-Net~\cite{chen2019automated} & \scriptsize B-Net~\cite{wu2017automatic} & \scriptsize PFA~\cite{wang2019multi} & \multicolumn{2}{c|}{\small VF~\cite{kim2020automation}} & \multicolumn{2}{c|}{\small Seg4Reg~\cite{lin2019seg4reg}} & \multicolumn{2}{c}{\small Ours} \\
        \hline
        Backbone & - & - & - & res50 & UNet & MNet & res18 & eff\_b1 & res18 & eff\_b1 \\
        \hline
        MAE & 8.58 & 10.46 & 9.31 & 6.69 & 3.90 & \textbf{3.51} & 6.63 & 3.96 & 4.50 & 3.73 \\
        \hline
        SMAPE (\%) & 20.35 & 26.94 & 23.44 & 12.97 & 8.79 & 7.84 & 10.95 & 7.64 & 8.47 & \textbf{7.32}\\
        \bottomrule
    \end{tabular}
    }
    \label{tab3}
\end{table}

\noindent\textbf{Comparison with State-of-the-art.}
We compare the proposed method with state-of-the-art methods including A-Net~\cite{chen2019automated}, L-Net~\cite{chen2019automated}, BoostNet~\cite{wu2017automatic}, PFA~\cite{wang2019multi}, VF~\cite{kim2020automation}, and Seg4Reg~\cite{lin2019seg4reg}, which won the 1st place in AASCE challenge. We follow the same experiment setting with VF~\cite{kim2020automation}, and the performance of all competing methods is adopted from the original publications for a fair comparison.

As depicted in Table~\ref{tab3}, promising results are observed in predicting Cobb angle using the proposed framework. The SMAPE is 7.32\% with our Seg4Reg+ framework, whereas PFA, VF-MNet and Seg4Reg achieve 12.97\%, 7.84\% and 7.64\%, respectively. The MAE of our method is 3.73, slightly inferior to VF-MNet. We believe the superior performance of our model in SMAPE is more convincing as SMAPE is the only evaluation metric in the AASCE challenge.

\section{Conclusion}
In this paper, we proposed a novel Seg4Reg+ model for automated Cobb angle estimation. Our Seg4Reg+ model incorporates segmentation into regression task via an attention regularization module and a region-of-interest enhancement gate to boost the performance of both tasks. Further, the two networks are integrated with global consistency learning for global optimization. Experimental results demonstrated the effectiveness of each module and the superior performance of our model to the state-of-the-art methods.

\noindent\textbf{Acknowledgments.} This work was funded by Key-Area Research and Development Program of Guangdong Province, China (No. 2018B010111001), and the Scientific and Technical Innovation 2030-"New Generation Artificial Intelligence" Project (No. 2020AAA0104100).

\bibliographystyle{splncs04}
\bibliography{ref}
\end{document}